\pdfoutput=1

\documentclass[11pt]{article}

\usepackage{additional_pkgs}

\usepackage[]{ACL2023}

\usepackage{times}
\usepackage{latexsym}

\usepackage[T2A, T1]{fontenc}

\usepackage[utf8]{inputenc}

\usepackage{microtype}

\usepackage{inconsolata}

\usepackage[bulgarian, greek, main=english]{babel}

\newcommand{\STAB}[1]{\begin{tabular}{@{}c@{}}#1\end{tabular}}

\title{Massively Multilingual Lexical Specialization of Multilingual Transformers}

\author{Tommaso Green\textsuperscript{1}, \textbf{Simone Paolo Ponzetto\textsuperscript{1}}, \textbf{Goran Glava\v{s}\textsuperscript{2}} \\
  \textsuperscript{1} Data and Web Science Group, University of Mannheim, Germany \\
  \textsuperscript{2} CAIDAS, University of Würzburg, Germany \\
  \texttt{\{tommaso.green, ponzetto\}@uni-mannheim.de}  \\ 
 \quad \texttt{goran.glavas@uni-wuerzburg.de}}

\newcommand\new{}

\definecolor{Gray}{gray}{0.92}
\newcolumntype{Y}{>{\centering\arraybackslash}X}

\renewcommand\labelenumi{(\roman{enumi})}
\renewcommand\theenumi\labelenumi

\begin{document}
\maketitle
\begin{abstract}

While pretrained language models (PLMs) primarily serve as general-purpose text encoders that can be fine-tuned for a wide variety of downstream tasks, recent work has shown that they can also be rewired to produce high-quality word representations (i.e., static word embeddings) and yield good performance in type-level lexical tasks. While existing work primarily focused on the lexical specialization of single monolingual PLMs, in this work we expose massively multilingual transformers (MMTs, e.g., mBERT or XLM-R) to multilingual lexical knowledge at scale, leveraging BabelNet as the readily available rich source of multilingual and cross-lingual type-level lexical knowledge. 
Concretely, we use BabelNet's multilingual synsets to create synonym pairs (or synonym-gloss pairs) across 50 languages and then subject the MMTs (mBERT and XLM-R) to a lexical specialization procedure guided by a contrastive objective. We show that such multilingual lexical specialization brings substantial gains in two standard cross-lingual lexical tasks, bilingual lexicon induction and cross-lingual word similarity, as well as in cross-lingual sentence retrieval. 
Crucially, we observe gains for languages unseen in specialization, indicating that multilingual lexical specialization enables generalization to languages with no lexical constraints. In a series of controlled experiments, we show that the number of specialization constraints plays a much greater role than the set of languages from which they originate.

\end{abstract}

\section{Introduction}
\label{sec:intro}

Massively multilingual transfomers (MMTs) such as mBERT \cite{devlin-etal-2019-bert} and XLM-R \cite{conneau-etal-2020-unsupervised}, among others, have been the primary vehicle of cross-lingual NLP transfer, offering state-of-the-art performance for many tasks and target languages in various zero-shot and few-shot transfer scenarios \cite[\textit{inter alia}]{pires-etal-2019-multilingual,wu-dredze-2019-beto,cao2020multilingual,artetxe-etal-2020-cross,lauscher-etal-2020-zero,zhao-etal-2021-closer,ruder-etal-2021-xtreme}. 
Much less work, however, investigated their capabilities as multilingual type-level word encoders \cite{vulic-etal-2020-probing}. Recent work, focusing primarily on monolingual PLMs, demonstrated that they can be turned into effective type-level lexical encoders using lexical constraints \cite{vulic-etal-2021-lexfit,liu-etal-2021-fast}, i.e., a process commonly referred to as \textit{lexical specialization}. Existing work, however, investigated lexical specialization in monolingual or bilingual settings only, namely specializing PLMs for a single language or a pair of languages using either English monolingual or noisy translated bilingual lexical constraints \cite{vulic-etal-2021-lexfit}. This is not only computationally inefficient, since one specialization needs to be executed for each language or language pair, but also does not tap into the wealth of multilingual knowledge of MMT's simultaneous pretraining on many (100+) languages, as well as large amounts of manually curated knowledge from massively multilingual knowledge bases like BabelNet \cite{navigli-ponzetto-2010-babelnet}.

In this work, in contrast, we investigate massively multilingual lexical specialization of MMTs, i.e., the potential benefits and limitations of a \textit{single lexical specialization procedure in multiple languages}. To this end, we tap into BabelNet as the readily available massively multilingual lexico-semantic resource. Concretely, we release a dataset of synonym pairs or synonym-gloss pairs that cover 50 languages (representing 14 different language families) obtained from BabelNet's multilingual synsets and leverage them as positive instances in a contrastive specialization training procedure. Our evaluation on two multilingual lexical tasks -- bilingual lexicon induction (BLI) and cross-lingual semantic word similarity (XLSIM) -- as well as on the task of cross-lingual sentence retrieval demonstrate the effectiveness of the multilingual lexical specialization when compared against vanilla MMTs.

We complement our evaluation with diagnostic experiments aimed at studying  properties of the multilingual lexical constraints that might drive the downstream lexical performance of the specialized models, namely the choice of the languages and the number of constraints. For this, we perform experiments where we control MMT specialization for (i) the linguistic diversity of the language represented in the specialization dataset and  (ii) the size of the specialization dataset, i.e., the number of synonym pairs from BabelNet used in contrastive training. 
The results of our diagnostic experiments suggest that, counterintuitively,  the typological diversity of the languages used in specialization (i.e., the specialization languages) has barely any effect in  defining the downstream performance. These findings for multilingual specialization for (type-level) lexical tasks contrast the observations for higher-level tasks, requiring the modeling of sentence or sentence-pair semantics, in which both multi-source specialization/fine-tuning on diverse languages \cite{chen-etal-2019-multi-source,ansell-etal-2021-mad-g} and linguistic proximity between training and evaluation languages \cite{lin-etal-2019-choosing,lauscher-etal-2020-zero} have been shown to strongly affect the transfer performance. At the same time, we find that the alignment performance quickly saturates with few constraints: this corroborates the \emph{rewiring hypothesis} of \citet{vulic-etal-2021-lexfit}, here in a  massively multilingual setting. To encourage further research on this topic, we release our code and all our datasets of lexical constraints and synonym-gloss pairs at \url{https://github.com/umanlp/babelbert}.  

\section{Related Work}
\label{sec:related_work}

External lexical knowledge from lexico-semantic resources (e.g., WordNet, ConceptNet) has been extensively leveraged for improving distributional representations of words -- a process commonly referred to as semantic specialization. Earlier work on semantic specialization of word embeddings can roughly be divided into  
(i) \textit{joint specialization} approaches \cite[inter alia]{yu-dredze-2014-improving, xu2014, bian2014, kiela-etal-2015-specializing, liu-etal-2015-learning, ono-etal-2015-word}, which integrate the external lexical constraints in the word embedding (pre)training and (ii) \textit{retrofitting} or \textit{postprocessing} techniques that post-hoc modify the pretrained embeddings, conforming them to external lexical constraints \cite{faruqui-etal-2015-retrofitting, wieting-etal-2015-paraphrase,mrksic-etal-2017-semantic,vulic-etal-2018-post,glavas-vulic-2018-explicit,ponti-etal-2018-adversarial}. 

External lexical knowledge has also been used to enrich monolingual PLMs by coupling masked language modeling with an auxiliary pretraining objective using lexical relations from WordNet  \cite{lauscher-etal-2020-specializing,levine-etal-2020-sensebert}.
However, these approaches are all aimed at enriching the Transformer with additional knowledge to be exploited in various downstream tasks, rather than producing better type-level word representations.

Several paradigms for obtaining semantically improved multilingual representation spaces have been proposed for static word embeddings: (i) \textit{align-and-specialize} \cite{mrksic-etal-2017-semantic} starts from mutually unaligned monolingual embedding spaces and both aligns them and semantically specializes for semantic similarity (as opposed to other types of semantic relatedness) by means of both multilingual (i.e., monolingual constraints for multiple languages) and cross-lingual lexical constraints; (ii) in \textit{cross-lingual specialization transfer} \cite{vulic-etal-2018-post,glavas-vulic-2018-explicit} the embedding space of a target language is first projected into the (unspecialized) space of the high-resource language (typically English) using a limited amount of cross-lingual source-target lexical alignments and then specialized with the specialization model trained using abundant monolingual lexical constraints of the resource-rich source language; (iii) \textit{constraint transfer} \cite{ponti-etal-2019-cross} noisily translates the (abundant) source language constraints into the target languages and uses those to specialize the monolingual embedding spaces of the target languages.

More recently, \citet{vulic-etal-2020-probing} showed that pretrained transformers encode a wealth of lexical knowledge in their parameters and that it is possible to obtain from them type-level (i.e., static, out-of-context) word representations that outperform representations obtained with word embedding models like fastText \cite{bojanowski-etal-2017-enriching}.
In a subsequent approach dubbed LexFit \cite{vulic-etal-2021-lexfit}, they specialize monolingual BERT models for a range of languages using English monolingual constraints from WordNet and Roget's Thesaurus, and automatically translated constraints \cite{ponti-etal-2019-cross} for languages other than English. Finally, once they obtain static word representations for each language, they induce a bilingual space in a standard fashion, by means of an orthogonal projection \cite{smith2017offline,artetxe-etal-2019-bilingual}. The effectiveness of LexFit stems from the fact that language-specific PLMs tend to produce better representations for a language than the MMTs (e.g., mBERT) due to the ``curse of multilinguality'' \cite{conneau-etal-2020-unsupervised,pfeiffer-etal-2022-lifting}. This, however, limits its applicability to a few languages with existing monolingual PLMs, crucially excluding low-resource ones. The approach is also computationally expensive as it entails (i) a separate specialization process for each language (plus a noisy translation of English constraints to the target language), and (ii) a bilingual alignment for each language pair. Additionally, it does not exploit cross-lingual lexical constraints in specialization, merely in post-hoc alignment.   
In this work, we propose a single massively multilingual lexical specialization approach (dubbed BabelBERT) to be applied to pretrained MMTs. To this end, we propose to obtain the lexical constraints from BabelNet, a massively multilingual lexico-semantic resource \cite{navigli-ponzetto-2010-babelnet,navigli2021ten}. BabelBERT has several advantages: (i) we leverage multilingual (i.e., monolingual from multiple languages) as well as \textit{cross-lingual} lexical constraints and (ii) perform a single multilingual specialization procedure, instead of specializing separately for each language or language pair; (iii) the multilingual specialization on top of MMTs removes the need for language-specific transformers and additionally allows for specialization effects to propagate to unseen languages, i.e., languages without readily available lexical constraints.         

\section{Multilingual Lexical Specialization}
\label{sec:methodology}

We first describe how we create lexical constraints from BabelNet in \S\ref{ssec:constraints} and then how we leverage them in a contrastive learning procedure in \S\ref{ssec:specialization}.  

\subsection{Constraint Mining}
\label{ssec:constraints}

In line with most work on lexical specialization for semantic similarity, we exploit the lexico-semantic relation of synonymy to obtain the specialization constraints. The multilingual synsets of BabelNet\footnote{We use version $5.0$, which covers $500$ languages, under the non-commercial license (\url{https://babelnet.org/full-license})} allow us to simultaneously obtain both monolingual and cross-lingual synonym pairs. 
Let $L = \{L_1, \dots, L_n\}$ be the predefined set of languages for which we mine the constraints and $W = \{w_1, \ldots, w_N\}$ a list of seed words. For each word, we first obtain all BabelNet synsets containing that word, discarding synsets that represent named entities and keeping only synsets that have at least two glosses in languages from $L$. We then iterate over all the fetched synsets, creating all possible (monolingual and cross-lingual) synonym pairs ($w_1$, $w_2$), each associated with a language pair ($L^{word}_1$, $L^{word}_2$). We also extract glosses ($g_1$, $g_2$), i.e., sentences that explain the concept of the synset, in languages $(L^{gloss}_1, L^{gloss}_2)$ different from $L^{word}_1$ and $L^{word}_2$, respectively.\footnote{$L^{word}_1$ may, however, be the same as $L^{gloss}_2$ and so may $L^{word}_2$ be the same as $L^{gloss}_1$.} To ensure the quality of the words we extract, we only keep a word if it lies in the top-$k$ words in its language frequency list.\footnote{We use the \textit{wordfreq} library \cite{wordfreq} and fastText vocabularies for the languages that \textit{wordfreq} does not cover.} We additionally discard words that are automatic translations or Wikipedia redirections, remove multi-words and delete duplicates.

\subsection{Specializing for Semantic Similarity}
\label{ssec:specialization}

Our lexico-semantic specialization procedure is based on a Bi-Encoder architecture (often also referred to as Dual-Encoder or Siamese architecture) and a contrastive training objective.

\paragraph{Type-Level Word Representations.} 
Following related work \cite{vulic-etal-2020-probing,bommasani-etal-2020-interpreting,vulic-etal-2021-lexfit}, we obtain type-level word representations from a PLM independently for each word from the synonym pair: we tokenize each word into its constituent subwords $sw_1 \dots sw_m$ and feed the sequence $[SPEC_{1}][sw_1]\dots[sw_m][SPEC_{2}]$ into the MMT, with $[SPEC_{1}]$ and $[SPEC_{2}]$ denoting the special sequence start and sequence end tokens of the MMT (e.g., $[CLS]$ and $[SEP]$, respectively, for BERT).
We get the final representation $\mathbf{e}_w^{type}$ by mean pooling the representations of its subwords (without the special tokens) from the last layer of the Transformer encoder.

\paragraph{Sense-Level Word Representations.} 
Type-level representations of polysemous words conflate, by construction, all of its senses into one embedding. To address this and make the representations capture a specific sense, we leverage additional context through sense-level information provided by the knowledge base. For example, among the senses of the word \textsf{bat} found in BabelNet, the list of the most frequent senses contains the synset for the nocturnal animal and the sports club (e.g., the one used for example in cricket). Besides the set of synonyms, i.e., the different synsets these senses belong to, their different meaning is captured by the glosses. The gloss for the animal sense reads as follows:
\begin{quote}
    Nocturnal mouselike mammal with forelimbs modified to form membranous wings [...]
\end{quote}
whereas for the sport stick we find: 
\begin{quote}
    A cricket bat is a specialised piece of equipment used by batters in the sport of cricket [...]
\end{quote}
In light of this, we make the MMT additionally encode word-gloss pairs, in order to obtain sense-level representations: to this end,
we append one of the mined glosses $g$ to the word as follows:  

\vspace{.5em}
\begin{center}
$[SPEC_{1}][sw_1]\dots[sw_m][SPEC_{2}] \, g \, [SPEC_{2}]$    
\end{center}
\vspace{.5em}

\noindent
and feed this as input to the MMT.
As with the type-level representations, we get the final representation $\mathbf{e}_w^{sense}$ by mean pooling the representations of \emph{only} the subwords $sw_1 \dots sw_m$ (i.e., gloss is there just for the contextualization of the word) from the last layer of the Transformer encoder.  

\paragraph{Contrastive Objective.} We train in batches $\mathcal{B} = {(w_1^{(i)}, w_2^{(i)}, \mathit{syn}^{(i)}_\mathit{id})}_{i = 1}^{N_B}$ of synonym pairs, with $\mathit{syn}^{(i)}_\mathit{id}$ denoting the BabelNet synset of the synonym pair $(w_1^{(i)}, w_2^{(i)})$.  In sense-level training, each data point additionally contains the glosses ($g_1^{(i)}$, $g_2^{(i)}$). We train by minimizing a variant of the popular InfoNCE contrastive loss \cite{infonce}. In a single batch, there might be more than one pair belonging to the same synset, we thus form all possible \emph{ordered}~\footnote{This is because the last $sim(\cdot)$ term depends on which word occupies the first position.} positive pairs in a set $\mathcal{P}$, i.e., pairs of words with the same $\mathit{syn}_\mathit{id}$.

\begin{align*}
\mathcal{L}_{\text{InfoNCE}}^{\mathcal{B}} = \frac{-1}{|\mathcal{P}|} \sum_{(w_1^{(i)}, w_2^{(j)}) \in \mathcal{P}}
 \log \left(\mathit{sim}(\mathbf{e}_{w_1}^{(i)}, \mathbf{e}_{w_2}^{(j)})\right) \\ - \log\left( \mathit{sim}(\mathbf{e}_{w_1}^{(i)}, \mathbf{e}_{w_2}^{(j)}) + \sum_{n \in \mathcal{N}^{(i)}} \mathit{sim}(\mathbf{e}_{w_1}^{(i)}, \mathbf{e}^{(n)}) \right)
\end{align*}
%
%
\noindent where $\mathit{sim}(\mathbf{e}_{w_1}^{(i)}, \mathbf{e}_{w_2}^{(j)}) = \exp (\cos(\mathbf{e}_{w_1}^{(i)}, \mathbf{e}_{w_2}^{(j)})) / \tau$, with $\tau$ as the temperature hyperparameter and $\mathcal{N}^{(i)}$ as the set of in-batch negatives, i.e., words from the other pairs in the batch that come from a BabelNet synset other than $\mathit{syn}^{(i)}_\mathit{id}$.

\paragraph{Adapter-Based Fine-Tuning.} 

Besides full fine-tuning of the MMT's parameters, we  experiment with lexical specialization via adapter-based fine-tuning~\citep{pmlr-v97-houlsby19a}. Adapters, shown useful in various sequential and transfer learning scenarios \cite{pfeiffer-etal-2020-mad,ruckle-etal-2020-multicqa,lauscher-etal-2021-sustainable-modular,hung-etal-2022-ds} are parameter-light modules that are inserted into a PLM's layers before specialization (i.e., fine-tuning): during specialization, only adapter parameters are tuned, while the PLM's pretrained parameters are kept fixed. We adopt the architecture of \newcite{pfeiffer-etal-2020-mad}, in which one bottleneck adapter is inserted into each Transformer layer.  

\section{Experimental setup}
\label{sec:experimental}

\paragraph{Multilingual Lexical Constraints.} 

We focus on the 50 diverse languages from the popular XTREME-R benchmark \cite{ruder-etal-2021-xtreme} which we report with language codes for brevity:
\textit{af}, \textit{ar}, \textit{az}, \textit{bg}, \textit{bn}, \textit{de}, \textit{el}, \textit{en}, \textit{es}, \textit{et}, \textit{eu}, \textit{fa}, \textit{fi}, \textit{fr}, \textit{gu}, \textit{he}, \textit{hi}, \textit{ht}, \textit{hu}, \textit{id}, \textit{it}, \textit{ja}, \textit{jv}, \textit{ka}, \textit{kk}, \textit{ko}, \textit{lt}, \textit{ml}, \textit{mr}, \textit{ms}, \textit{my}, \textit{nl}, \textit{pa}, \textit{pl}, \textit{pt}, \textit{qu}, \textit{ro}, \textit{ru}, \textit{sw}, \textit{ta}, \textit{te}, \textit{th}, \textit{tl}, \textit{tr}, \textit{uk}, \textit{ur}, \textit{vi}, \textit{wo}, \textit{yo}, \textit{zh}. The sample covers 14 language families (Afro-Asiatic,
Austro-Asiatic, Austronesian, Dravidian, Indo-
European, Japonic, Kartvelian, Kra-Dai, Niger-
Congo, Sino-Tibetan, Turkic, Uralic, Creole, and
Quechuan) and additionally contains Basque and Korean as two language isolates.

We collect the constraints from BabelNet with the procedure described in \S\ref{ssec:constraints}. As seed words, we select the top-$N$ English most frequent words ($N = 1,000$, filtering for stopwords using NLTK \cite{bird-loper-2004-nltk}) and retain only words that belong to the top-$k$ ($k=15,000$) words in the frequency list of a language. The total training set consists of 761,273 lexical constraints: we provide additional statistics of the dataset in appendix \ref{sec:train_stats} and a few examples in Table \ref{tab:babel-ex}.

\paragraph{Evaluation Tasks.} We evaluate on two standard cross-lingual word-level tasks, bilingual lexicon induction (BLI) and cross-lingual word similarity (XLSIM). We couple this with an evaluation on unsupervised cross-lingual sentence retrieval. For the word-level tasks (XLSIM and BLI): a) we make sure not to include in the training set of lexical constraints from BabelNet any word that appears in the test sets; b) we take the mean-pooling of the embeddings of the subwords from the best-performing layer (see Table \ref{tab:hyperparameters}) of the MMT as word representations.

\paragraph{Task 1: Bilingual Lexicon Induction.} BLI tests the quality of a multilingual (bilingual) representation space by means of type-level word alignment between languages. For a given query word from a source language, words from the vocabulary of a target language are ranked based on their similarity with the query. The position of the correct translation of the query in the target language ranking reflects the quality of the type-level word alignment between the languages.   
We evaluate on two well-established benchmarks:  
G-BLI \cite{glavas-etal-2019-properly} covers 28 language pairs between 8 languages
(\textit{de}, \textit{en}, \textit{fi}, \textit{fr}, \textit{hr}, \textit{it}, \textit{ru}, \textit{tr}), 
 whereas PanlexBLI \cite{vulic-etal-2019-really} spans 15 diverse languages
(\textit{bg}, \textit{ca}, \textit{eo}, \textit{et}, \textit{eu}, \textit{fi}, \textit{he}, \textit{hu}, \textit{id}, \textit{ka}, \textit{ko}, \textit{lt}, \textit{no}, \textit{th}, \textit{tr})
for a total of 210 language pairs. For both datasets, we evaluate on the test portions consisting of 2,000 word pairs per language pair and report the performance in terms of Mean Reciprocal Rank (MRR) as recommended by \citet{glavas-etal-2019-properly}. Following \citet{vulic-etal-2020-probing},
the vocabularies used for retrieval cover the top 100K most frequent words from the respective fastText Wikipedia vectors \cite{bojanowski-etal-2017-enriching}. 

\paragraph{Task 2: Cross-Lingual Word Similarity.} XLSIM measures the correlation between the similarities of cross-lingual word pairs obtained based on their representations in the multilingual (bilingual) representation space and similarity scores assigned by human annotators. We evaluate the performance on 66 language pairs between 12 languages 
(\textit{zh}, \textit{cy}, \textit{en}, \textit{et}, \textit{fi}, \textit{fr}, \textit{he}, \textit{pl}, \textit{ru}, \textit{es}, \textit{sw}, \textit{yue}) 
from the MultiSimLex dataset \cite{vulic-etal-2020-multi} and use Spearman's $\rho$ between the cosine similarities between words' embeddings and the corresponding human-assigned similarity scores.

\paragraph{Task 3: Cross-Lingual Sentence Retrieval.} For cross-lingual sentence retrieval, we use the Tatoeba dataset \cite{artetxe-schwenk-2019-massively} which comprises 112 languages, where each language has 1,000 sentences paired with their translations in English. We obtain the sentence embedding by mean-pooling the representations of all of its subword tokens at the output of the best-performing Transformer layer. We straightforwardly compute the similarities between sentences as the cosine of the angle between their embeddings. We compare each query sentence to its nearest neighbour and compute accuracy as our evaluation measure.

\paragraph{Training Details.}

We experiment using two different MMTs: multilingual BERT (mBERT) \cite{devlin-etal-2019-bert} and XLM-R \cite{conneau-etal-2020-unsupervised}.\footnote{We use \texttt{bert-base-multilingual-uncased} and \texttt{xlm- roberta-base} from  HuggingFace \cite{wolf-etal-2020-transformers}.} 
In all experiments, we set the temperature of the InfoNCE loss to $\tau= 0.07$. 

To account for the very skewed distribution of constraints across language pairs (cf.\ Figure \ref{fig:constraints}), following \citet{xlm}, we sample batch constraints from the multinomial distribution $\{ q_{i,j} \}$ over language pairs $(L_i,L_j)$ as follows:
\begin{equation*}
    q_{i,j} = \frac{p_{i,j}^{\alpha}}{\sum_{k,l} p_{k,l}^{\alpha}}, \quad p_{i,j} = \frac{n_{i,j}}{\sum_{k,l} n_{k,l}}
\end{equation*}
where $n_{i,j}$ denotes the number of synonym pairs for a language pair $(L_i, L_j)$ and $\alpha$ is the smoothing factor. We set $\alpha$ to $0.5$.

For model selection (both hyperparameter search and checkpoint selection), we proceed as follows. We randomly select two language pairs from G-BLI and two language pairs from PL-BLI and pick the corresponding training sets -- consisting of 5,000 pairs -- as our validation set. Before training, we run one validation loop to get the MRR score of the unspecialized vanilla MMT for each of the language pairs. During training, we stop every quarter of an epoch to perform validation: we track for each of these four validation language pairs the relative improvement of MRR w.r.t.\,the vanilla score. We use the average of these four relative improvements as our overall validation metric that guides model selection. 
We train for 15 epochs using AdamW \cite{adamw} and use PytorchLightning \cite{pl_lightning} for our implementation, coupled with Huggingface Transformers \cite{wolf-etal-2020-transformers} and Pytorch Metric Learning \cite{pytorchmetriclearning} libraries. For the adapter-based models, we use the adapter-transformers library \cite{pfeiffer-etal-2020-adapterhub}.

\paragraph{Hyperparameters and training details}
\label{sec:hyperparameters}
\new{For the fully fine-tuned models, we search for the optimal learning rate $lr \in \{2e-5,5e-6,1e-6\}$ and the batch size $N_{B} \in \{32, 64\}$. For adapter-based models, we additionally try $lr = 1e-4$ and set the adapter reduction ratio to $16$ (i.e., we set the bottleneck size of the adapters to $48$). Every experiment is done on a single NVIDIA V100 or A100 GPU on a computing cluster at our disposal. Taking into account failed experiments, grid searches and successful runs we report 331 days of compute (including CPU time for preprocessing and retrieval) as logged by the Weights \& Biases logger \cite{wandb}. We provide the full list of hyperparameter values in the Appendix (Table \ref{tab:hyperparameters}).}

\setlength{\tabcolsep}{2pt}
\begin{table}[t!]
\centering
\normalsize{
\begin{tabularx}{\linewidth}{llcccc}
\toprule
MMT & Model & \multicolumn{2}{c}{BLI} & XLSIM & Ttb \\ 
\cmidrule(lr){3-4}
    &       & G-BLI & PL-BLI      &    &            \\ \midrule
\multirow{4}{*}{\STAB{\rotatebox[origin=c]{90}{mBERT}}} & \cellcolor{Gray}{vanilla} & \cellcolor{Gray}{14.5} & \cellcolor{Gray}{10.7} & \cellcolor{Gray}{10.3} & \cellcolor{Gray}{34.1} \\
& Babel-Ad & 20.0 & 12.3 & 25.4 & 43.2 \\
& Babel-FT & \textbf{20.9} & \textbf{12.4} & \textbf{25.8} & \textbf{43.7} \\
& Babel-Gl & 19.5 & 11.8 & 24.1 & 41.7 \\
 \midrule
\multirow{4}{*}{\STAB{\rotatebox[origin=c]{90}{XLM-R}}} & \cellcolor{Gray}{vanilla} & \cellcolor{Gray}{8.5} & \cellcolor{Gray}{5.4} & \cellcolor{Gray}{5.9} & \cellcolor{Gray}{37.6} \\ 
& Babel-Ad & 17.8 & 8.7 & 32.0 & 55.6 \\
& Babel-FT & \textbf{20.8} & \textbf{10.5} & 34.2 & 55.7 \\
& Babel-Gl & 19.7 & 10.0 & \textbf{34.4} & \textbf{58.6} \\
\bottomrule
\end{tabularx}%
}
\caption{Results of multilingual lexical specialization for two MMTs -- mBERT and XLM-R on three tasks (four datasets): BLI -- on G-BLI and PanLex-BLI (PL-BLI), XLSIM, and cross-lingual sentence retrieval on the Tatoeba dataset (Ttb). Performance reported in terms of MRR ($\times100$) for BLI, Spearman's $\rho$ for XLSIM and accuracy ($\times100$) for Ttb. For each dataset, we report averages across all language pairs (28 for G-BLI, 210 for PL-BLI, 66 for XLSIM, and 112 for Ttb). The highest scores per column and MMT are in \textbf{bold}.}
\label{tab:main-res}
\end{table}

\section{Results and Discussion}
\label{sec:results}

We present our main results in Table~\ref{tab:main-res}, where we compare the multilingual lexical specialization procedure with type-level representations using full fine-tuning (Babel-FT) and adapter-based tuning (Babel-Ad) and full-fine tuning of the MMT using sense-level representations (Babel-Gl) against the baseline models provided by the unspecialized vanilla MMTs. We show the results for the word representations that come from the layer for which the best average performance is obtained on the given dataset: we provide the information on optimal layers for lexical representations in Table \ref{tab:hyperparameters} in the Appendix. For each model, we compute each language-pair score as the average over $3$ runs with different random seeds.

\begin{figure*}
    \centering
    \includegraphics[width=\textwidth]{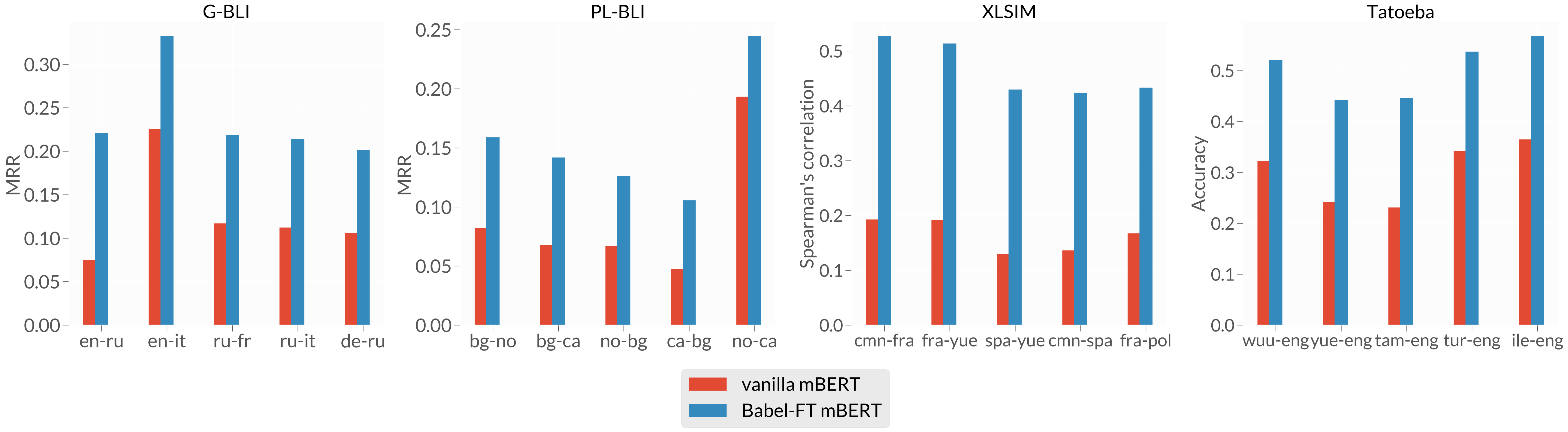}
    \caption{The five language pairs for each evaluation task/dataset for which we observe the largest performance improvement with Babel-FT mBERT. Each language pair score is obtained using the best layer for the specific dataset-model combination as reported in Table \ref{tab:hyperparameters}.}
    \label{fig:top-k-langs}
\end{figure*}

Overall, the results indicate that multilingual lexical fine-tuning improves the performance of both MMTs (mBERT and XLM-R) on all tasks. All three specialization variants (Babel-Ad/FT/Gl) yield similar performance on all three tasks for mBERT. The same is, however, not the case for XLM-R, where full fine-tuning (Babel-FT and Babel-Gl) leads to substantially better performance across the board compared to adapter-based training (Babel-Ad). Training with sense-level information in the form of synset glosses (Babel-Gl) seems particularly beneficial for cross-lingual sentence retrieval on Tatoeba (Ttb) -- we assume that this is because, much like the rest of the sentences in Tatoeba, the glosses in specialization training provide sentential context to word representations. 
Interestingly, despite the fact that vanilla mBERT produces substantially higher quality word representations than vanilla XLM-R (e.g., 14.5 vs. 8.5 on G-BLI or 10.3 vs. 5.9 on XLSIM), our multilingual lexical specialization tilts the results in favour of XLM-R, with comparable BLI results between the two and much better XLM-R performance on XLSIM (9-point gap) and Ttb (15-point gap). 
This suggests that XLM-R actually contains richer multilingual lexical knowledge than mBERT, which is, however, buried deeper in the model (this is corroborated by the fact that for XLM-R we generally get better lexical representations from lower Transformer layers, see the Appendix): once uncovered by means of our multilingual lexical specialization, this richer lexical information of XLM-R surfaces.

Figure \ref{fig:top-k-langs} shows the five language pairs for each evaluation task/dataset for which we observe the largest performance improvement with Babel-FT mBERT. 
\new{In PL-BLI, we find two languages, Norwegian and Catalan, for which no constraint was seen during training: this indicates that the benefits of massively multilingual lexical specialization propagate to unseen languages. What we find particularly encouraging is the fact that Tatoeba pairs with the largest gains include some low-resource languages and dialects (e.g., Wu and Yu Chinese, Tamil, and Interlingue), some of them not even present in MMTs in pretraining.}

\subsection{Additional Analysis}
\label{ssec:further_analysis}

\setlength{\tabcolsep}{4.7pt}
\begin{table*}[!t]
\centering
\small{
\begin{tabular}{p{7cm}rrr}
\toprule
\multicolumn{1}{l}{Language sample} & $d_{\text{typ}}$ & {$sim_{train-test}$} &  PL-BLI Avg  \\ \midrule
\{af, de, el, en, es, fr, nl, pt, ro, uk\} & 0.1830 & 0.7369 & 0.1154 \\
\{af, en, hi, it, ms, nl, pt, ro, ru, uk\} & 0.2406 & 0.7275 & 0.1176 \\
\{af, de, en, fr, it, pl, pt, ta, uk, ur\} & 0.2661 & 0.7113 & 0.1164 \\
\{bn, de, en, es, hi, it, jv, ro, ru, uk\} & 0.2906 & 0.7114 & 0.1159 \\
\{en, it, mr, ms, pl, pt, ru, ta, uk, zh\} & 0.3180 & 0.6953 & 0.1166\\
\{az, de, en, fa, fr, it, pt, sw, ta, uk\} & 0.3576 & 0.6764 & 0.1172 \\
\{af, de, el, fr, jv, ml, mr, qu, ta, uk\} & 0.3746 & 0.6520 & 0.1176 \\
\{bn, el, fr, hi, jv, mr, ms, qu, ro, tl\} & 0.4022 & 0.6530 & 0.1175 \\
\{bn, gu, it, my, pt, ro, sw, ta, vi, zh\} & 0.4231 & 0.6345 & 0.1162 \\
\{ar, de, el, en, ja, jv, kk, my, sw, ur\} & 0.4600 & 0.6207 & 0.1171 \\
\bottomrule
\end{tabular}
} 
\caption{Selection of $10$ samples of $10$ languages each in increasing $d_{\text{typ}}$ order, together with the similarity between training and test languages and average MRR on PL-BLI.}
\label{tab:typ_div_exp}
\end{table*}

We perform additional ablations to try to isolate the factors that lead to performance gains from massively multilingual lexical specialization. To this end, we carry out  experiments in which we vary (i) the typological diversity of the sample of languages from which we take BabelNet constraints and (ii) the size of the training set of lexical constraints used for specialization.

\paragraph{Role of Linguistic Diversity.} We investigate how changing typological diversity of the sample of languages from which we draw the specialization constraints affects the generalization to unseen languages. That is, we test whether a selection of languages with different degrees of linguistic diversity and no overlap with the test languages impacts the multilingual lexical specialization of the MMTs. To quantify linguistic diversity, we borrow the typological diversity index $d_{\text{typ}}$ from \citet{ponti-etal-2020-xcopa}. For a sample of languages $S$, we compute the index based on the URIEL vectors \cite{littell-etal-2017-uriel} of languages in the sample.\footnote{We use the \texttt{syntax\_knn} vectors which contain $103$ manually-coded features indicating various syntactic properties of languages.} We obtain $d_{\text{typ}}$ of the sample $S$ by computing an entropy value for each of the features across all languages in $S$: such value is 0 if all languages have identical values for that feature. We then  average the entropy scores across all features. 

We choose PL-BLI as our benchmark for this analysis as it proved to be the most challenging lexical task. We first create 10 samples, each containing 10 languages, as follows: we first sample 1 million different language samples of 10 languages, making sure none of them contains any of the PL-BLI test languages. We then divide them into 10 bins according to $d_{\text{typ}}$ and randomly pick one sample from each bin at random. For each sample, we mine the synonym pairs from BabelNet from scratch, considering only those languages, and making sure that each language pair has exactly 100 constraints -- resulting in a total of 5,500 instances for each sample (counting both cross-lingual and monolingual constraints).
 In an effort to limit test language leakage in the training procedure, differently from the main experiments, we only validate on two language pairs from G-BLI for model selection. We conduct experiments with Babel-FT mBERT as our best-performing model on PL-BLI. 
 We fine-tune for 10 epochs with the same hyperparameters used in the main experiments but without the upsampling  as all language pairs have the same number of constraints.  We report the results, together with the language samples and their $d_{\text{typ}}$ score in Table \ref{tab:typ_div_exp}. Surprisingly, we observe a poor correlation between $d_{\text{typ}}$ of the language sample and the corresponding PL-BLI performance.

\begin{figure*}
    \centering
    \includegraphics[scale=0.31]{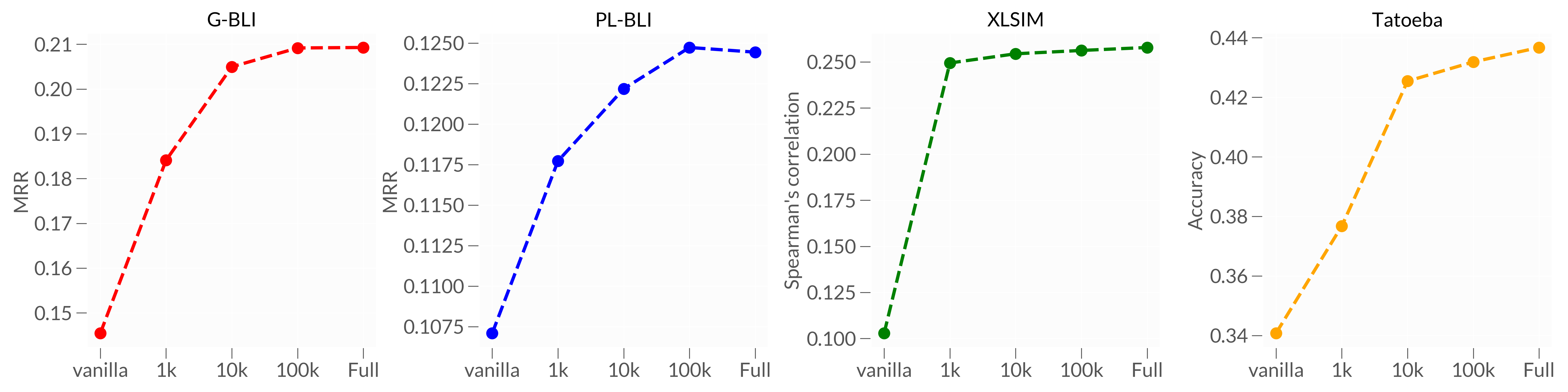}
    \caption{Performance on the four datasets plotted against the size of the specialization dataset: vanilla refers to the unspecialized MMT and Full to the full training set size (761,273).}
    \label{fig:constraint-analysis}
\end{figure*}

We additionally quantify the degree of similarity between the languages of each sample and the languages included in PL-BLI: $sim_{train-test}$ (also shown in Table \ref{tab:typ_div_exp}) is the average of pairwise similarities of URIEL vectors between the languages of the two sets.  

We again observe a poor correlation between $sim_{train-test}$ and PL-BLI performance, indicating that the proximity of training and testing languages does not play a role. While counterintuitive, this is actually favourable: it indicates that we can, with similar specialization success, leverage constraints from a wide range of languages, which allows us to mine them for high-resource languages for which structured lexical knowledge is more abundant.

\paragraph{The Role of Constraint Size.} We perform additional experiments to investigate the effect of training set size (i.e., the total number of constraints). For this, we create three additional training sets of sizes 1K, 10K, and 100K instances, with the same relative distribution of constraints across language pairs as in the full training. 
We then subject mBERT to Babel-FT specialization for 10 epochs (same hyperparameters and training procedure  as in the main evaluation). 
For each training set, we execute 3 different runs and report averages in Figure \ref{fig:constraint-analysis}. On all four benchmarks, we observe the same overall behaviour: the performance saturates already with 10K constraints. Tatoeba, the only sentence-level task in our setup, seems to be the task that benefits the most from a larger training set. In contrast, XLSIM performance saturates already with 1K lexical constraints, which seems to be in line with findings of \citet{vulic-etal-2021-lexfit}.

\citet{vulic-etal-2021-lexfit} proposed and empirically validated the \textit{rewiring hypothesis} -- that lexical specialization primarily exposes knowledge that is already present in the pre-trained weights, rather than injecting new knowledge. We believe our results confirm this claim: if specialization mainly resurfaces lexical knowledge hidden in the weights, this puts a cap on the downstream performance. In our case, we find that such knowledge seems to be independent of both the typological diversity of the training languages and the typological similarity of training and test languages: in this sense, the MMT appears to be learning a language-agnostic lexical alignment function that affects its entire representation space. Moreover, learning of this function only seems to require about 10K samples, with performance quickly saturating with more constraints.

\section{Conclusion}
\label{sec:conclusion}

In this work, we presented a multilingual lexical specialization approach that leverages the massively multilingual lexical knowledge available in BabelNet.  Differently from prior approaches, which perform monolingual or bilingual specialization procedures, we subject MMTs to a single training regime with lexical constraints from 50 languages and report substantial gains over the unspecialized baseline (i.e., MMT not subject to lexical specialization). We perform both type-level lexical specialization, i.e. with words fed in isolation to the transformer, and sense-level lexical specialization, by accompanying each word with a gloss. 
We perform a series of additional experiments to study the driving factors of lexical specialization: in one experiment, we keep the training set size fixed while diversifying the languages in the training set. We observe that this does not seem to have a significant impact on the overall performance. In a subsequent experiment, we again use constraints encompassing 50 languages, but limit the number of constraints per language pair: we find that more constraints help the model perform better, however, few samples are necessary to reach peak performance. Our results support the rewiring hypothesis of \citet{vulic-etal-2021-lexfit} that lexical specialization resurfaces the existing lexical knowledge stored in MMTs, rather than injecting it. Such extraction seems to be independent of the training languages and quickly saturates with few constraints. 

\paragraph{Limitations.}
While we try to perform multilingual lexical specialization on a set of typologically diverse languages, we are still restricting our analysis to a small fraction of all the languages of the world. In addition to this, our analysis investigates only two MMTs -- albeit arguably the two most widely used. Due to hardware limitations, we experimented with XLM-R Base: the results we report may be substantially different for XLM-R Large (or other larger MMTs like mT5), which possibly encodes more lexical knowledge.

\paragraph{Ethical considerations.}
We leverage lexical constraints from BabelNet, a resource constructed semi-automatically. BabelNet may contain lexical associations reflecting negative social biases (e.g., sexism or racism). Biased  constraints, when used as training data in our specialization, may strengthen societal biases present in MMTs.

\paragraph{Acknowledgments}

Tommaso Green and Simone Ponzetto have been supported by the JOIN-T 2 project of the Deutsche Forschungsgemeinschaft (DFG). Goran Glava\v{s} has been supported by the EUINACTION grant funded by NORFACE Governance (462-19-010) through Deutsche Forschungsgemeinschaft (DFG; GL950/2-1). We additionally acknowledge support by the state of Baden-Württemberg through bwHPC and the German Research Foundation (DFG) through grant INST 35/1597-1 FUGG. We thank our colleague Sotaro Takeshita for insightful discussions during the development of this project and Ines Rehbein for her comments on a draft of this paper.

\bibliography{anthology,custom}
\bibliographystyle{acl_natbib}

\appendix
\section{Training set statistics}
\label{sec:train_stats}

\begin{figure*}[t!] 
    \makebox[\textwidth][c]{\includegraphics[width=1.25\textwidth]{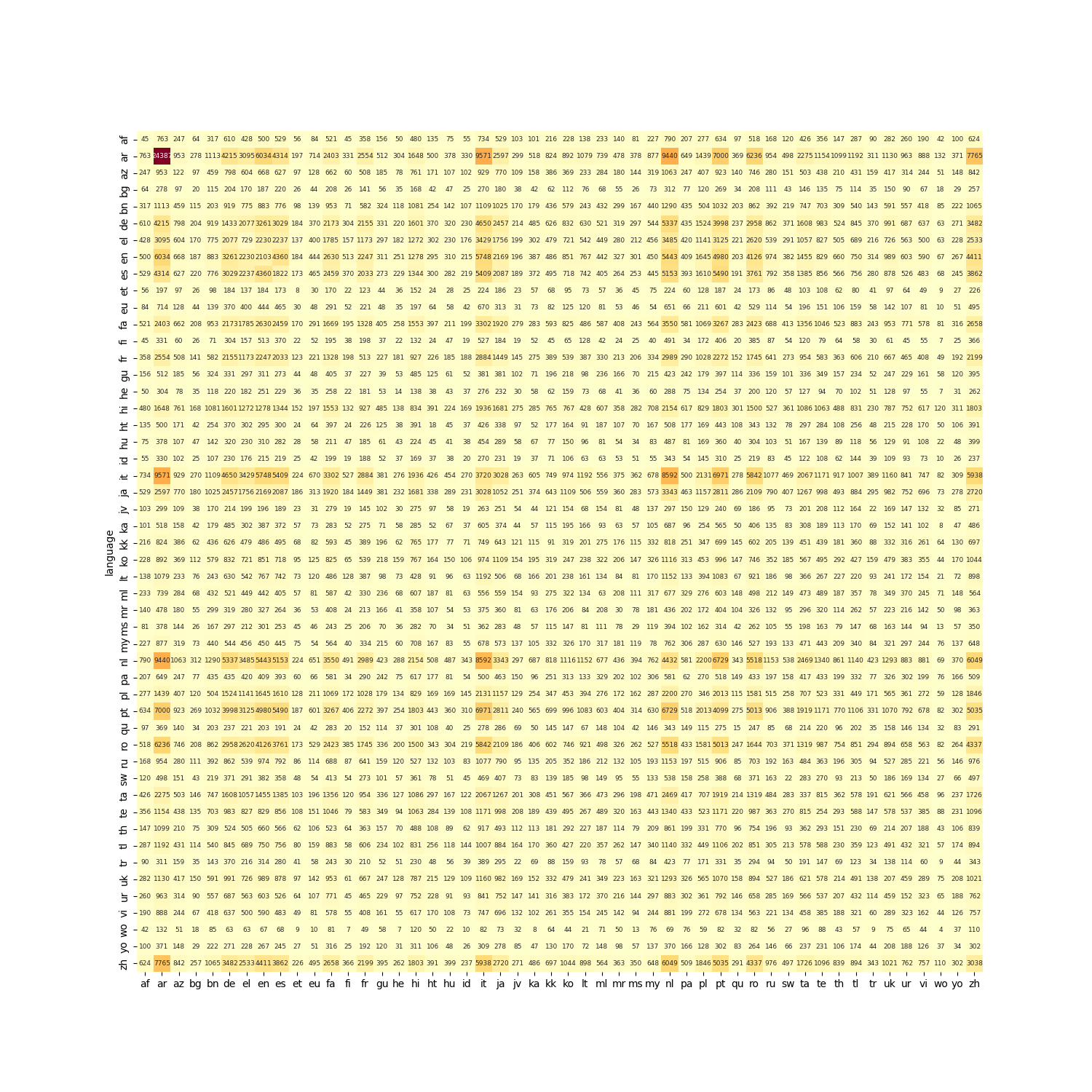}}%
    \caption{The numbers of (monolingual and cross-lingual) synonymy constraints in the training set for the selection of $50$ languages from the XTREME-R benchmark.}
    \label{fig:constraints}
\end{figure*}

We report the number of constraints for each language pair in Figure \ref{fig:constraints}. 


\setlength{\tabcolsep}{2pt}
\begin{table*}[]
\def\arraystretch{1.05}
\footnotesize
\raggedleft
\begin{tabularx}{\textwidth}{llXXl}
\toprule
\multicolumn{1}{c}{$w_1$} & \multicolumn{1}{c}{$w_2$} & \multicolumn{1}{c}{$g_1$} & \multicolumn{1}{c}{$g_2$} & 
\multicolumn{1}{c}{$syn_{id}$} \\
\midrule
production {[}en{]} &
  produit {[}fr{]} &
  \foreignlanguage{bulgarian}{Нещо, произведено от човешка или механична дейност или при естествен процес.} {[}bg{]} &
  The amount of an artifact that has been created by someone or some process. {[}en{]} &
  bn:00064584n \\
passato {[}it{]} &
  gestern {[}de{]} &
  \foreignlanguage{bulgarian}{Минало е период от време, поредица от събития, които вече са се случили.} {[}bg{]} &
  Iragana edo lehena gauden unearen aurretik gertatu den oro dugu. {[}eu{]} &
  bn:00060927n \\
  lajan {[}ht{]} &
  centen {[}nl{]} &
  \foreignlanguage{greek}{Το χρήμα είναι οποιοδήποτε στοιχείο ή επαληθεύσιμη εγγραφή που είναι γενικά αποδεκτό ως πληρωμή για αγαθά} [...] {[}el{]} &
  Montant non spécifié d'une devise. {[}fr{]} &
  bn:00055644n \\
lungsod {[}tl{]} &
  oraș {[}ro{]} &
  A large, busy city, especially as  the main city in an area  or country or as distinguished from surrounding rural areas. [en] &
  Una metropoli è una città  di grandi dimensioni  con più di 1 milione  di abitanti con un'area comunale {[}...{]} {[}it{]}  & bn:00019319n \\
\bottomrule
\end{tabularx}%
\caption{Examples of synonym pairs together with their glosses and synset identifiers.}
\label{tab:babel-ex}
\end{table*}

\begin{table*}[t!]
\centering
\small{
\begin{tabularx}{\linewidth}{XXXXXXXX}
\toprule
MMT & Model & {$lr$} & {$N_B$} & {G-BLI Layer} & {PL-BLI Layer} & {XLSIM Layer} & {Ttb Layer} \\ 
\midrule
\multirow{4}{*}{mBERT} & vanilla & - & - & 6 & 6 & 6 & 8  \\
& Babel-Ad & $5e-6$ & $32$ & $12$ & $12$ & $12$ & $11$  \\
& Babel-FT & $1e-6$ & $64$ & $12$ & $12$ & $12$ & $11$  \\ 
& Babel-FT-Gl & $1e-6$ & $64$ & $12$ & $12$ & $12$ & $11$ \\
\midrule
\multirow{4}{*}{XLM-R} & vanilla & - & - & $1$ & $1$ & $1$ & $7$  \\
& Babel-Ad & $1e-4$ & $64$ & $8$ & $8$ & $12$ & $10$  \\
& Babel-FT & $1e-6$ & $32$ & $11$ & $12$ & $12$ & $11$  \\ 
& Babel-FT-Gl & $5e-6$ & $32$ & $9$ & $8$ & $12$ & $10$\\
\bottomrule
\end{tabularx}%
}
\caption{The best-found hyperparameters of the models together with the index of the best-performing layer for each dataset.}
\label{tab:hyperparameters}
\end{table*}

\end{document}